 \documentclass[pmlr,twocolumn, tablecaption=bottom]{jmlr} 

\usepackage{booktabs}
\usepackage[load-configurations=version-1]{siunitx} 
 
\usepackage{microtype}
\usepackage{graphicx}
\usepackage{xcolor}
\usepackage{amsmath,amssymb,scrextend}
\usepackage{bm}
\usepackage{bbm}
\usepackage{mathtools}
\usepackage{xparse}
\usepackage{comment}

\DeclarePairedDelimiterXPP\Expect[2]{\mathbb{E}_{#1}}[]{}{#2}%

\DeclareMathOperator*{\btheta}{\bm{\theta}}

\theorembodyfont{\upshape}
\theoremheaderfont{\scshape}
\theorempostheader{:}
\theoremsep{\newline}

\jmlrvolume{LEAVE UNSET}
\jmlryear{2020}
\jmlrsubmitted{LEAVE UNSET}
\jmlrpublished{LEAVE UNSET}
\jmlrworkshop{Machine Learning for Health (ML4H) 2020} 

\title[Trust Issues: Uncertainty Estimation Does Not Enable Reliable OOD Detection]{Trust Issues: Uncertainty Estimation Does Not Enable Reliable OOD Detection On Medical Tabular Data}

 \author{%
   \Name{Dennis Ulmer} \Email{dennis.ulmer@mailbox.org}\\
   \Name{Lotta Meijerink} \Email{lotta.meijerink@pacmed.nl}\\
   \Name{Giovanni Cinà} \Email{giovanni.cina@pacmed.nl}\\
   \addr Pacmed BV - Amsterdam, Netherlands\\
 }

\begin{document}

\maketitle

\begin{abstract}
    When deploying machine learning models in high-stakes real-world environments such as health care, it is crucial to accurately assess the uncertainty concerning a model's prediction on abnormal inputs. However, there is a scarcity of literature analyzing this problem on medical data, especially on mixed-type tabular data such as Electronic Health Records.
    We close this gap by presenting a series of tests including a large variety of contemporary uncertainty estimation techniques, in order to determine whether they are able to identify out-of-distribution (OOD) patients. In contrast to previous work, we design tests on realistic and clinically relevant OOD groups, and run experiments on real-world medical data. We find that almost all techniques fail to achieve convincing results, partly disagreeing with earlier findings. 
\end{abstract}

\section{Introduction}

In healthcare, tremendous potential has been identified for applications deep learning methods for e.g. prediction, screening, finding or designing better treatments for patients \citep{sheikhalishahi2019benchmarking, esteva2019guide}. However, safe deployment in practice demands certain properties from models: in the context of healthcare, models have to prove to be interpretable and deliver trustworthy predictions  \citep{he2019practical}. In particular, a trustworthy application should be able to identify new samples outside the training population. Although we always hope for models to generalize to novel data points, a significant shift in the data can render the current predictor unreliable. The COVID-19 pandemic offers an example of this phenomenon on a large scale: the training data used for models deployed before the outbreak might not have contained patients with these new combinations of symptoms. Thus, the predictions of said models on COVID-patients should have a lower degree of certainty to avoid misleading end-users like doctors and nurses. The same could happen due to other, more common factors such as changes in the patient demographics, evolving clinical protocols or scaling of a device \citep{curth2019transferring}.

Deep neural networks are notorious for delivering predictions in an overconfident manner, be it by confidently predicting the wrong class after minimal perturbation of an input \citep{goodfellow2015explaining}, assigning high probability to OOD samples \citep{nalisnick2018deep} or simply being badly-calibrated in general \citep{guo2017calibration}. Such observations have motivated the development of a series of techniques to more accurately estimate the uncertainty about an outcome \citep{gal2016dropout, lakshminarayanan2017simple, blundell2015weight}.  Alas, there is still a limited volume of experiments on tabular data and classification tasks, a context of particular importance for health care applications. Notably, \citet{ovadia2019can} conducted an extensive study of the quality of uncertainty estimates under covariate shift, by artificially constructing corrupted OOD samples. We argue that this procedure is insufficient to evaluate these techniques in a real-world healthcare application: the generalization of performance to OOD examples in an artificial scenario is no guarantee of robustness, especially if the model in question is unable to detect OOD reliably. Inability to recognize OOD could result in confident predictions on abnormal data points for which the model actually under-performs. 

We study the efficacy of a large array of uncertainty estimation methods for OOD detection in the context of a binary classification task with clinical significance, i.e. prediction of mortality for ICU patients, by running experiments on two large mixed-type, highly unbalanced tabular data sets. 
We provide three main contributions.\footnote{All code is publicly available under \url{https://github.com/Pacmed/ehr_ood_detection}.} First, we design a number of clinically relevant OOD experiments, including scenarios for changing patient population, variation of clinical protocol and data corruption. Second, we show that all tested uncertainty estimation techniques for neural discriminators fail to reliably detect OOD samples across experiments. Lastly, we provide novel benchmarks for OOD detection on two large tabular medical data sets.

\section{Related Work}

Uncertainty from a modelling perspective is often separated into aleatoric and epistemic uncertainty. While the former usually refers to \emph{non-reducible} uncertainty, e.g. uncertainty inherent in the data generating process, the latter denotes \emph{reducible} uncertainty. This encompasses both the lack of knowledge about the ideal model for a problem, as well as about its best parameters \citep{der2009aleatory, hullermeier2019aleatoric}. In our case, we are concerned with methods that try to estimate either only the epistemic or both types of uncertainty in a post-hoc manner. 

One method to appraise uncertainty estimation methods is the study of their behaviour in presence of OOD samples \citep{ovadia2019can}. 
The topic of OOD detection has been studied extensively; we present here a selection of articles. A simple baseline for discriminators for the task was introduced by \citet{hendrycks2017baseline}.  \citet{lee2018simple} fit a generative model on the pre-softmax outputs of neural discriminators and employ the Mahalanobis distance of data samples as a confidence metric. \citet{jiang2018trust} employ the distance between a sample and the next sample of another class as a non-probabilistic trust score. Other articles use density estimation models instead \citep{xiao2020likelihood, choi2018waic}, while some identify their shortcomings when it comes to detecting OOD samples in the image domain \citep{nalisnick2018deep, kirichenko2020normalizing}.
\citet{eduardo2020robust} introduce a variant of the variational autoencoder for tabular data, which is able to assign OOD scores to both single cells (i.e. features) and row entries (i.e. whole data points) alike.
A related line of work investigates the phenomenon of covariate shift: \citet{ovadia2019can} conduct a large study to analyze model calibration under increasing shift. \citet{park2020calibrated} propose a procedure to explicitly adapt models to shifting data distributions. There is also some work in the medical context about this topic, with e.g. \citet{curth2019transferring} deriving formal guarantees for domain adaptation procedures and testing them on clinical data sets.

Given the importance of the reliability of models in a health care setting, the application of such methods to tabular medical data sets constitutes important related work. 
However, the literature on the topic is scarce. \citet{myers2020identifying} introduce a model-independent reliability metric based on the difference between the model's prediction and predictions for patients of the same class. 
\citet{ruhe2019bayesian} and \citet{meijerink2020uncertainty} test different models for mortality prediction on electronic health records, albeit with limited experiments, while \citet{dusenberry2020analyzing} also test models and their uncertainty in predicting diagnoses. The latter work however restricts its scope to only two types of models and does not evaluate uncertainty estimates for novel patient groups.

\section{Background}\label{sec:background}

First of all, we lay out some relevant definitions and notations. Let $p: \mathbb{R}^D \rightarrow [0, 1]$ denote a probability density function from which data samples $\vec{x}_i \in \mathbb{R}^D$ stem from and let $p(\vec{x}, y): \mathbb{R}^D\times\{0, 1\} \rightarrow [0, 1]$ denote the joint distribution of points and labels with $y \in \{0, 1\}$ in a binary classification setting. We indicate with $\mathcal{D}$ the set of tuples contained in a data set.
We consider a data point out-of-distribution if its underlying distribution $q$ underwent a \emph{covariate shift} relative to our model's training distribution $p$, namely if $p(\vec{x})$ and $q(\vec{x})$ are different.\footnote{Other authors such as \citet{shimodaira2000improving} and \citet{moreno2012unifying} give more restrictive definitions of covariate shift (e.g. assuming $p(y|\vec{x}) = q(y|\vec{x})$) that do not apply in our setting.} Covariate shift is especially prevalent in non-stationary environments \citep{moreno2012unifying}, which abound in healthcare \citep{curth2019transferring}. 

We consider the following metrics for OOD detection: the maximum softmax probability baseline by \citet{hendrycks2017baseline} and the standard deviation of probabilities of class $1$ aggregated from multiple different sources, e.g. different members of an ensembles or predictions produced under distinct sets of weights samples from a posterior distribution. In these cases, we  also compute the predictive entropy \citep{gal2016uncertainty} or mutual information  \citep{smith2018understanding} across different predictions. Due to space limitations, we refer the reader to Appendix \ref{app:uncertainty-metrics} for a more in-depth explanation of these methods.

\section{Methodology}

We  describe the experimental setup, the models employed and the evaluation metrics.

\subsection{Experimental Design}\label{sec:data}

\paragraph{Mortality Prediction Task} As we are interested in testing the OOD-detection capabilities of models trained on a clinically relevant classification task, we focused only on prediction of in-hospital mortality based on the first $48$ hours of data from intensive care admissions. This task has already been investigated extensively as a relevant task for clinical prediction models \citep{sheikhalishahi2019benchmarking, ruhe2019bayesian, curth2019transferring, harutyunyan2019multitask, meijerink2020uncertainty} and provides a clear example of binary classification with unbalanced outcome.

\paragraph{Clinical Data Sets} We performed experiments on the MIMIC-III data set \citep{johnson2016mimic}, which comprises health data from ICU admissions from the Beth Israel Deaconess Medical Center in Boston, Massachusetts. From the $46476$ patients and $61532$ ICU stays in the MIMIC-III data set, data points are selected based on patient age ($\ge 18$ years) and length of stay ($\ge 48$ hours). Furthermore, stays were excluded when there was no data in the first $48$ hours or when there were multiple IC transfers within one hospital admission. After selection, the cohort consisted of $21139$ IC stays corresponding to $18094$ patients, with a mortality rate of $13.23 \%$. We also employed the eICU data set \citep{pollard2018eicu}, which contains ICU data from patients admitted to different hospitals in the United States in 2014 and 2015. We pre-processed the data using the general pipeline presented in the work of \cite{sheikhalishahi2019benchmarking} and applied the same filtering criteria used for MIMIC-III, resulting in $38072$ ICU stays, with one unique patient per stay and a mortality rate of $12.53 \%$.

\paragraph{Feature Engineering} To simplify our experimental pipeline and allow more direct comparisons between the two data sets, we only keep clinical variables that were present in both data sets (see appendix Table \ref{tab:variable-names}). We then engineered features calculating six statistics on seven sub-sequences of a time series, in line with the feature engineering for the logistic regression model of \citet{harutyunyan2019multitask}. The sub-sequences consist of the full time series, the first $10 \%$ and last $10 \%$, the first and last $25 \%$, and the first and last $50 \%$. The statistics are minimum, maximum, mean, standard deviation, skew and the number of measurements, which were all standard-scaled and mean-imputed if necessary, resulting in $588$ different features. In both cases, the data sets were split into 70 $\%$ training, 15 $\%$  validation and 15 $\%$ test set.

\paragraph{Clinically Relevant OOD Groups}
Using the help of medical professionals, we select clinically relevant patient groups on both datasets.  
We first extract groups of patients based on age (newborns for MIMIC-III\footnote{We were unable to identify enough newborns matching our filter criteria in eICU.}), on ethnicity (white, black) and on gender, to test the model's behavior under changing patient demographics. We also separate patients by admission type (elective vs. emergency admissions) to simulate change in protocol and according to diagnosis (acute and unspecified renal failure, epilepsy, hypertension with complications and secondary hypertension and thyroid disorders) as we also envision novel sets of symptoms to emerge in patients in practice. To ensure that the resulting patients groups are sufficiently different, we perform a feature-wise Welch's t-test and report the percentage of features that differ in a statistically significant manner from the rest of the training set; the percentages are reported in the plots of the results in  Section \ref{sec:results}.

\subsection{Models}\label{sec:models}

In the following, we briefly discuss the models and corresponding OOD detection metrics used for our experiments. Due to space constraints we refer to the cited papers for a more detailed description of the models.

\paragraph{Single Models} We consider different variations of a standard feed-forward neural network with a final sigmoid activation function. Besides a normal network with ReLU activation functions and intermediate dropout layers, we also consider a model which was additionally calibrated with temperature scaling \citep{guo2017calibration} on the validation set. As uncertainty metrics, the model's entropy over the class distribution as well as its maximum softmax probability are considered. We also add a logistic regression baseline.

\paragraph{Bayesian Models} We include models using MC Dropout \citep{gal2016dropout} and Bayes-by-Backprop (BBB, \citeauthor{blundell2015weight}, \citeyear{blundell2015weight}). As uncertainty metrics, we consider the standard deviation, their predictive entropy as well as their mutual information. We also attempted to train Gaussian Processes as a Bayesian baseline, but it could not scale to the size of the data set and did not converge to useful solutions on subsets.

\paragraph{Ensemble Models} We furthermore investigate the performance of different ensemble models, namely a group of standard feed-forward neural networks \citep{lakshminarayanan2017simple}, a bootstrapped ensemble, meaning that every member is trained on a different (but not disjoint) subset of the data, as well as the anchored bayesian ensembles introduced by \citet{pearce2020uncertainty}. The same metrics are employed as for the bayesian models.

\paragraph{Density Estimation Baselines} Lastly, we train two simple density estimation models: Probabilistic principal component analysis (PPCA; \citeauthor{bishop1999bayesian}, \citeyear{bishop1999bayesian}), where we simply use the log-probability of a sample as an indication of its novelty, as well as an auto-encoder (AE). For the latter, its reconstruction error serves as uncertainty metric, since we expect the model to reconstruct samples closer to the training distribution in a more reliable way. Although AEs are not explicitly estimating the data density, there is a known connection to PPCA as they project samples into a latent space, albeit in a non-linear fashion (cp. e.g. \citeauthor{valpola2015neural}, \citeyear{valpola2015neural}).

\subsection{Model Training \& Evaluation}

Random hyperparameter search \citep{bergstra2012random} was performed for all model types on MIMIC-III and eICU separately. The options considered, number of trials and best found hyperparameters are reported in Appendix \ref{app:hyperparams}. All models were generally trained for at most $10$ epochs using the adam optimizer \citep{kingma2014adam} and dropout \citep{srivastava2014dropout} regularization, with early stopping if the validation loss did not decrease after $3$ epochs. Logistic regression was trained using $l_2$ regularization.

To evaluate all the models enumerated in the last section, we consider the area under the receiver operating characteristic curve (AUC-ROC). 
In the case of mortality prediction, AUC-ROC is used as a standard classification metric. In the case of OOD detection, AUC-ROC is calculated for a classification task in which the classes are in-distribution and out-of-distribution. Here, the model is not evaluated based on its predicted class probabilities (about a patient's mortality), but rather on the uncertainty about its prediction. In short, AUC-ROC is employed to measure whether uncertainty helps in discriminating OOD samples.

\section{Results}\label{sec:results}

In this section we describe the main experimental findings on mortality prediction and OOD detection.

\subsection{Mortality Prediction}\label{sec:mortality-prediction}

Before delving deeper into OOD detection, we compare the models' performances on the mortality prediction task, the results of which are given in Table \ref{tab:mortality prediction}. Almost all models are able to solve the task well, achieving very similar results across data sets. Neural models slightly outperform the logistic regression baseline. One exception is the anchored ensemble, which seems to perform slightly worse. We attribute this effect to the regularization procedure using sampled anchored points, which might prevent ensemble members to converge to the same local minimum. This seems to constitute a trade-off between the bayesian guarantees and performance, which will become relevant again in the following section. 
Furthermore, BBB performs considerably worse than the other models while also displaying a comparatively high standard deviation in results. We observed this even after an extensive hyperparameter search (for details see Appendix \ref{app:hyperparams}). We note that previous work applied BBB to similar data sets, but either using less features \citep{ruhe2019bayesian} or more homogeneous ones (MNIST in the case of \citeauthor{blundell2015weight}, \citeyear{blundell2015weight}); we hypothesize that a mixture prior of only two gaussian distributions for the weights might not be expressive enough to find a suitable weight distribution for the high number and complexity of features we employ.\footnote{The sensitivity of variational methods - including Bayes-by-Backprop - towards the choice of prior is part of an ongoing discussion, refer e.g. to the blog post by \citet{gelada2020bayesian}.}

\begin{table}[ht]
    \floatconts
        {tab:mortality prediction}%
        {\caption{\footnotesize Mortality prediction AUC-ROC on MIMIC-III and eICU. Results are averaged over $n=5$ runs.}}%
        {
            \resizebox{\columnwidth}{!}{%
            \begin{tabular}{lll}
                \toprule
                Model &              MIMIC &               eICU \\
                \midrule
                \texttt{AnchoredNNEnsemble}     &  $0.837 \pm 0.006$ &  $0.832 \pm 0.004$ \\
                \texttt{BBB}                    &  $0.628 \pm 0.081$ &  $0.611 \pm 0.066$ \\
                \texttt{BootstrappedNNEnsemble} &  $0.847 \pm 0.000$ &  $0.847 \pm 0.000$ \\
                \texttt{LogReg}                 &  $0.835 \pm 0.000$ &  $0.823 \pm 0.000$ \\
                \texttt{MCDropout}              &  $0.849 \pm 0.002$ &  $0.844 \pm 0.001$ \\
                \texttt{NNEnsemble}             &  $0.847 \pm 0.000$ &  $0.847 \pm 0.000$ \\
                \texttt{NN}                     &  $0.847 \pm 0.002$ &  $0.842 \pm 0.002$ \\
                \texttt{PlattScalingNN}         &  $0.847 \pm 0.002$ &  $0.844 \pm 0.002$ \\
                \bottomrule
            \end{tabular}%
            }
        }
    \label{tab:mortality prediction}
\end{table}

\subsection{OOD Experiments}

\begin{figure*}[ht]
    \floatconts
        {fig:perturbation_eicu}%
        {\caption{\footnotesize  Perturbation experiment results for eICU measured via the OOD detection AUC-ROC. Scales are given on the y-axis, tested models and metrics on the x-axis. Results are averaged over $n=100$ different, randomly selected perturbed features.}}%
        {\includegraphics[width=2\columnwidth]{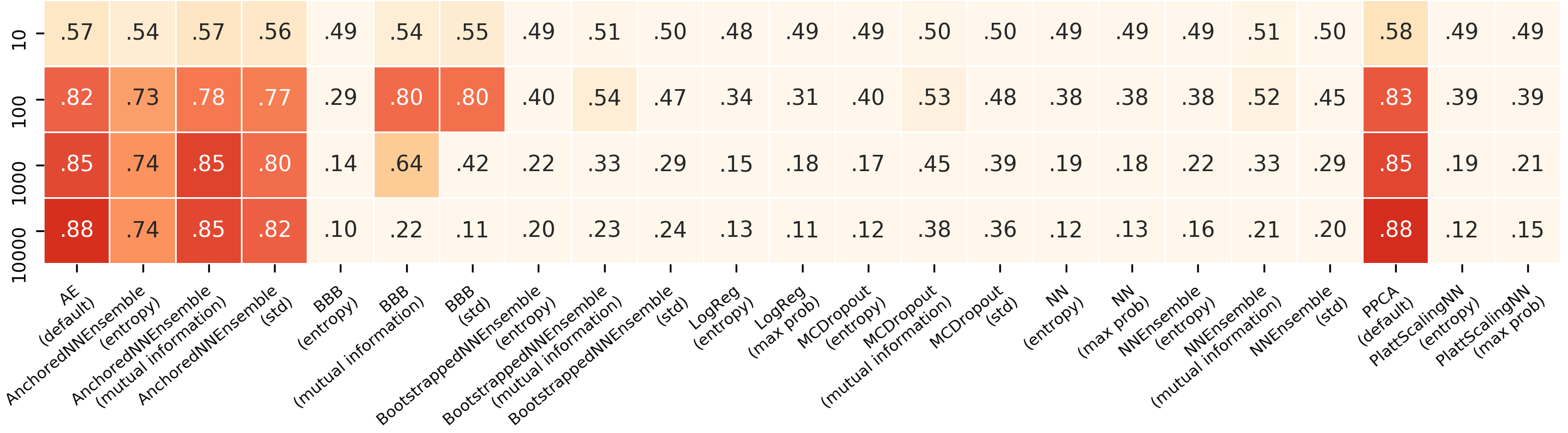}}
\end{figure*}

We conduct a series of experiments to determine whether the presented uncertainty estimation techniques are able to successfully detect OOD data. In a clinical setting, a change in the patient population is often more complex than the shift in a single feature, but it rather appears as a new configuration of covariates.  A model may encounter OOD examples for a variety of reasons: a change in protocol that modifies the provenance of the patients, an intrinsic shift in the patient population, or corruption in the data due to machine or human error, to name a few. To simulate these scenarios we confront our models with novel data obtained via three methods. First, data corruption: we shift the data artificially, scaling one feature at a time by an increasingly large degree. Second, changing patient population: some clinically relevant patient groups (based on demographics, provenance or pathology), are excluded from the training set and reintroduced later for testing (such as in \citeauthor{meijerink2020uncertainty}, \citeyear{meijerink2020uncertainty}). Third, changing data source: we repeat the same procedure, but this time using patient stays from eICU as OOD data for models trained on MIMIC-III, and vice versa.

\subsubsection{Artificial Data Corruption}\label{sec:data-corruption}

We investigate the ability of models to identify novel data points with abnormal feature values. To simulate extreme or possibly corrupted feature values, we scale features by a factor of $10$, $100$, $1000$ and $10000$. This is achieved by selecting a single feature at random and scaling it by the chosen factor across the whole test set. The corrupted test set and the original one are then compared in terms of uncertainty using AUC-ROC, to check whether uncertainty helps in identifying OODs. For each factor, we repeat this procedure one hundred times per model, sampling without replacement the feature to scale. We then report the average OOD detection AUC-ROC score on these samples, which is depicted for eICU in Figure \ref{fig:perturbation_eicu}; due to their similarity, the results for MIMIC-III were moved to Figure \ref{fig:perturbation_mimic} in the appendix.
In this case, we would like to see the AUC-ROC increase as the scaling of a random feature increases, i.e. the covariate of a sample becomes more and more abnormal compared to the training distribution. However, we see quite the opposite in most models regardless of the applied uncertainty estimation technique, as AUC-ROC actually \emph{decreases}, resulting in scores much worse than random guessing. In practice, this would result in a model being overly confident in a prediction for a corrupt data point. The only notable exceptions to this trend are AE and PPCA, which can be sorted into an entirely different family of methods and, surprisingly, the anchored ensemble and BBB. 

\subsubsection{Clinical OOD Groups}\label{sec:clinical-cohorts}

\begin{figure*}[ht]
    \floatconts
        {fig:ood_mimic}%
        {\caption{\footnotesize OOD AUC-ROC score for different models and uncertainty metrics for groups in the MIMIC-III data set. \emph{size} denoted the relative size of the OOD group compared to the full training set, \emph{diff} the percentage of OOD features that were different in a statistically significant manner compared to the remaining training set, using a Welch's t-test and $p < 0.01$. Results are averaged over $n=5$ runs. }}%
        {\includegraphics[width=2\columnwidth]{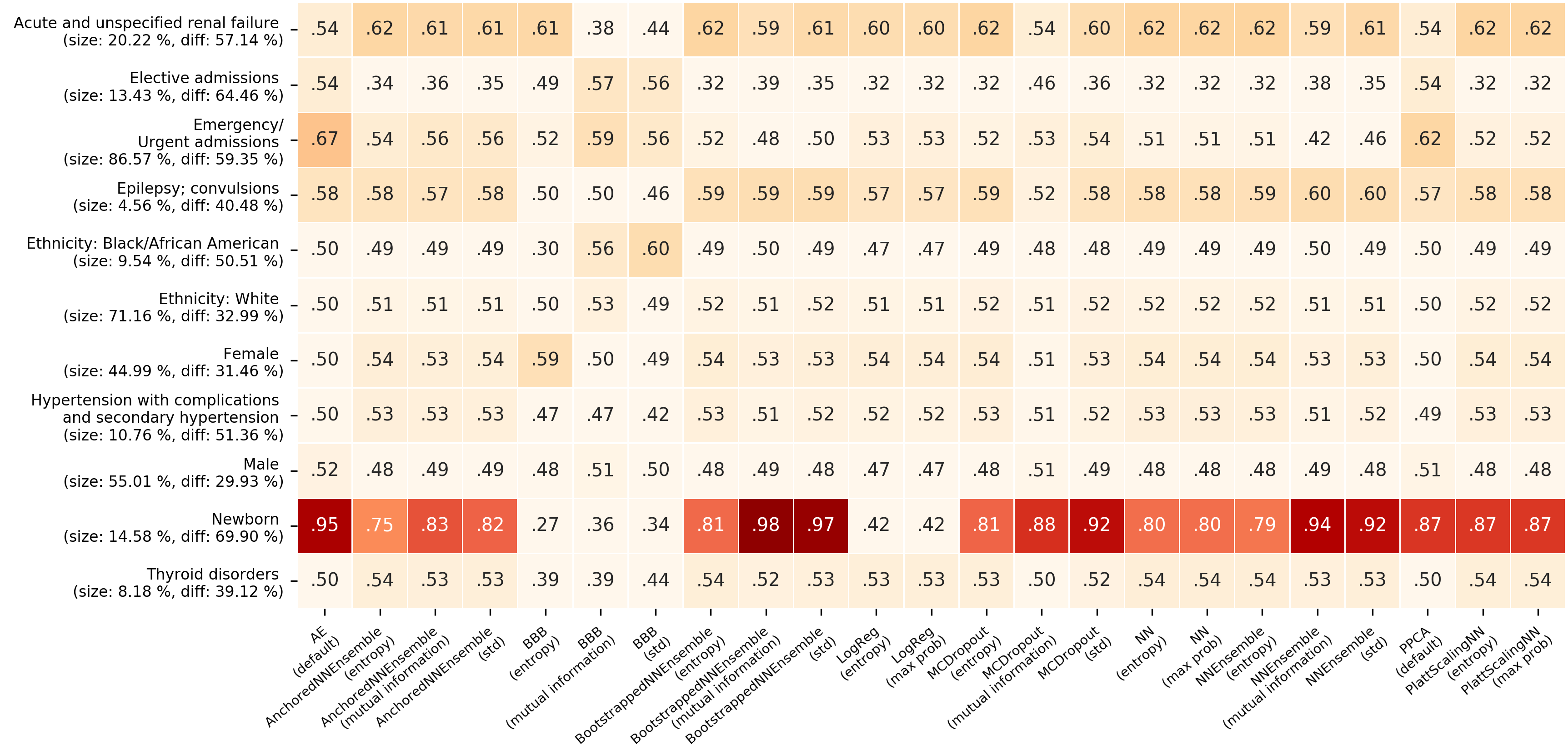}}
\end{figure*}

Several articles testing OOD detection capabilities employ images from other data sets as OOD samples (e.g. \citeauthor{pearce2020uncertainty}, \citeyear{pearce2020uncertainty} or  \citeauthor{kirichenko2020normalizing}, \citeyear{kirichenko2020normalizing}). In these cases, the OOD property is constituted by a disjunct set of class labels, e.g. when models are trained on MNIST digits but tested on clothing items. We deem this to be an inapplicable methodology for our predictive task, as it is very unlikely that new classes would be introduced in a clinical binary classification problem. We thus use the clinically relevant groups defined earlier, separate them from the training set and treat them as OOD during inference, staying within the same classification framework.

The results in Figure \ref{fig:ood_mimic} paint a discouraging picture:\footnote{Results for eICU are displayed in Figure \ref{fig:ood_eicu} in the appendix, while mortality prediction AUC-ROC scores for the OOD groups are given in Figure \ref{fig:ood-both-mortality}.} except for newborns in the MIMIC-III data sets, which have the highest percentage of significantly different features among all OOD groups, all models fail to discern in-distribution from out-of-distribution samples. This even holds for AE and PPCA. The difference between the model families is that density estimation models do seem to struggle for many groups, but do not fail as spectacularly as discriminators (see e.g. the 
``Elective admissions'' group in Figure \ref{fig:ood_mimic}).

\subsubsection{Clinical data sets as OOD}\label{sec:clinical-data-sets}

\begin{figure*}[ht]
    \floatconts
        {fig:ood-da}%
        {\caption{\footnotesize OOD detection AUC-ROC for models trained on eICU and tested on MIMIC-III (first row) and vice versa (second row). Results are averaged over $n=5$ runs.}}%
        {\includegraphics[width=2\columnwidth]{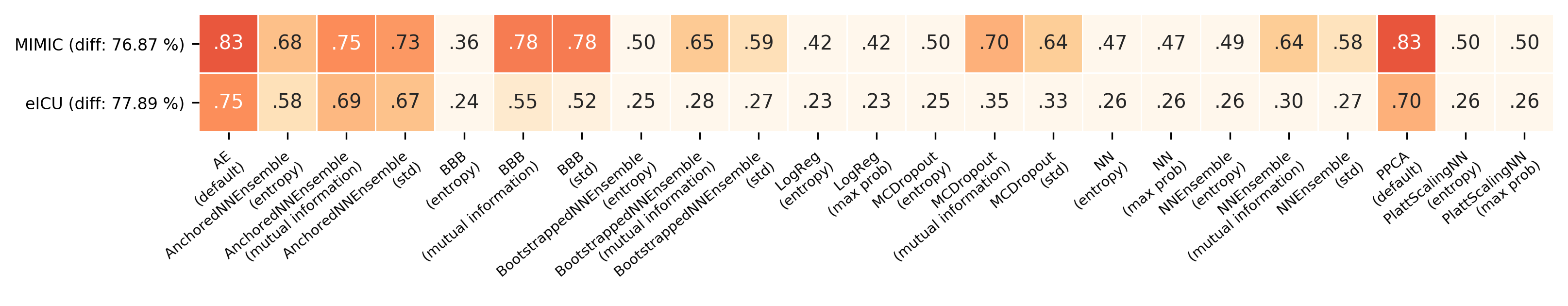}}
\end{figure*}

Lastly, we treat the whole of eICU as OOD examples for models trained on MIMIC-III and vice versa. This is different from a domain adaptation problem for mortality prediction, where we test the generalization of model performance on data from a different source.\footnote{These results are given in tab. \ref{tab:da-mortality} in the appendix.} We are instead interested to see whether models can identify the data as having a different origin. In the medical domain, this is especially relevant, as e.g. changing hospital protocols could have impact on patients' covariates and therefore influence a model's prediction.

The results in Figure \ref{fig:ood-da} illustrate the intricacies of the problem: again, the density estimation baselines perform best in both cases. By plotting both data sets in the same feature space, it becomes apparent that they are almost completely disjoint, exemplifying a clear case where $p(\vec{x}) \neq q(\vec{x})$ (see Figure \ref{fig:eicu-mimic-manifold} in the appendix). Furthermore, this fact also underlines the importance of this issue for medical applications: models trained on either data set might produce unreliable results for patients originating from a different source, whence the need to carefully assess the scaling of medical devices. In our case, classification performance generalizes well across both data sets (see Table \ref{fig:ood-both-mortality} in the appendix), however, there are documented cases in the literature where this is not the case \citep{curth2019transferring}.  There are some notable exceptions in the MIMIC-III to eICU case, namely BBB and anchored ensemble
, but also some slightly improved scores in the case of the bootstrapped ensemble and the MC Dropout model. Nonetheless, the fact that these scores cannot be reproduced in the reverse case - eICU to MIMIC-III - shows that these results cannot be seen as reliable but rather due to the idiosyncrasies of the training data and contingent model weights. 

\section{Discussion}

For several deep neural discriminators, we have demonstrated that the tested uncertainty estimation methods do not detect OOD samples reliably on medical tabular data. The intuition behind this shortcoming can be explained by the toy example in Figure \ref{fig:single_uncertainty}: By separating the feature space into potentially open-ended decision regions, neural discriminators will create large areas with high confidence where no training data were observed, making them prone to failure when presented with OOD samples. Every method aggregating multiple predictions will only enlarge the areas with low confidence due to the overlap of slightly different decision boundaries, but is unlikely to remove this effect entirely and reliably. We did see a few methods stand out in some experiments, but they were not able to do so consistently. In particular, the BBB and anchored ensemble performed comparatively well in the perturbation experiments. 

For the latter model, we conjecture that this is due to the special regularization, namely the $l_2$ distance to anchor points, which are sampled in the beginning of the training. This seems to incentivize the ensemble members to diversify,\footnote{Such a trade-off between diversity and regularization has also been noted by the original authors in a follow-up work \citep{pearce2018bayesian}.} an effect that we could also observe when comparing the predictive entropy scores for normal and anchored ensembles on toy data
(cp. Figure \ref{fig:compare-ensembles} in the appendix). When different models arrive at diverse solutions, their decision regions have less overlap, creating areas where combined discordant predictions produce higher uncertainty. 
A similar effect might be at play for BBB, although the diversity that seems to enable the model to identify OODs is more likely due to underfitting.
\footnote{BBB generally struggled to converge during training, even under various hyperparameter configurations (see Section \ref{sec:results} for further discussion).}

\begin{figure}[ht!]
    \floatconts
        {fig:single_uncertainty}%
        {\caption{\footnotesize Predictive entropy of a single neural classifier for a toy multi-class classification problem.}} %
        {\includegraphics[width=0.85\columnwidth]{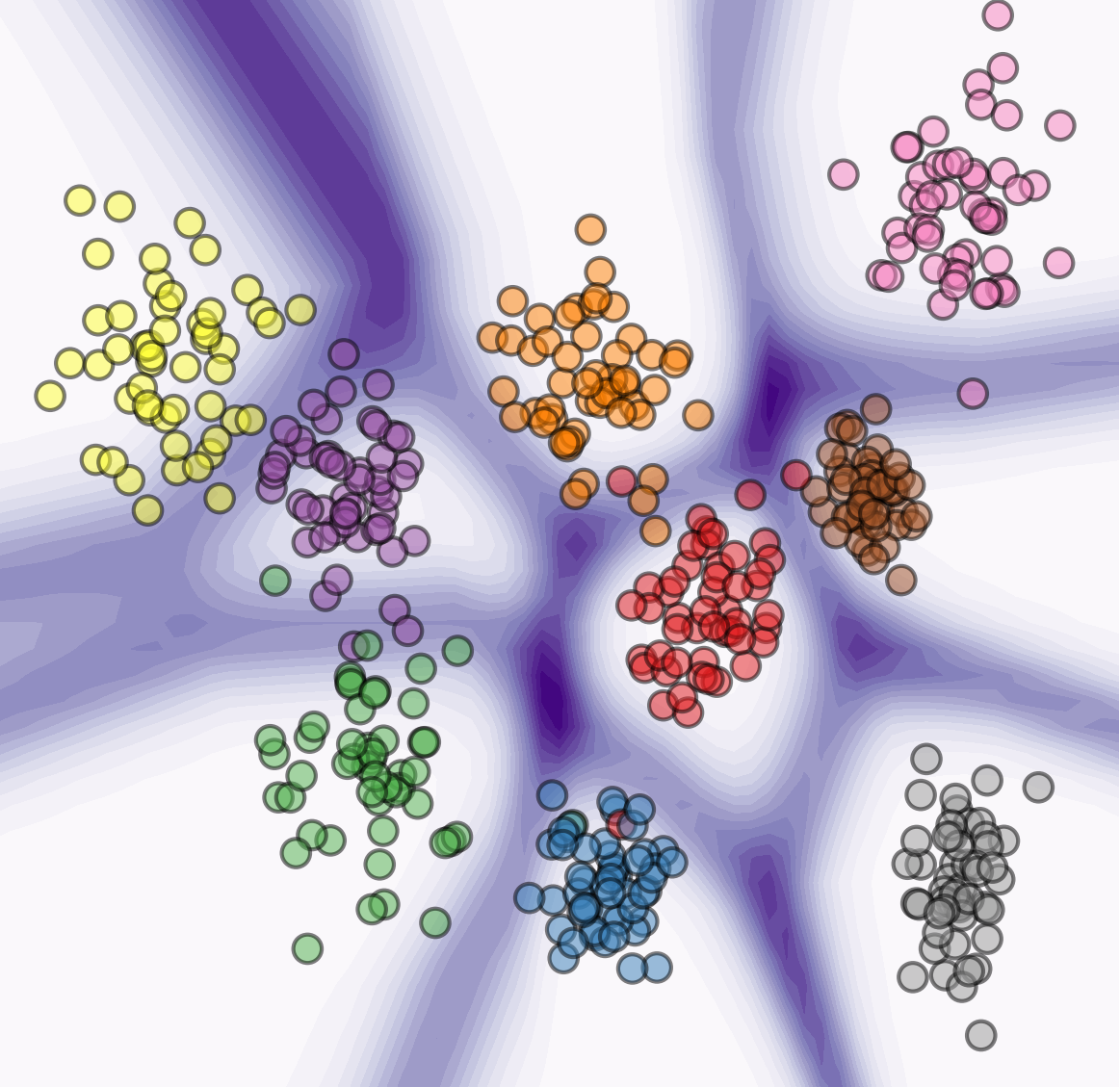}}
\end{figure}

For simple density-estimation models, many clinically-relevant patient groups also seem to be hard to identify as OOD. This may be because said groups appear to be interspersed with the remaining training population, even if a lot of their feature values differ significantly (cp. Figure \ref{fig:eicu_elective_admissions} in the appendix for the 
``elective admissions'' group in the eICU data set). An explanation  could be that the ability to distinguish samples probably does not only depend on the number of differently distributed features, but also on the features themselves, as models are likely learning to prioritize some features over others in terms of importance. More powerful density-estimation models might be able to distinguish these patients; it is an exciting direction for future work to assess whether models such as the one presented in \citet{eduardo2020robust}, which are tailored towards tabular data, or hybrid models, e.g. in the work of \citet{grathwohl2020classifier}, can achieve better OOD detection on such patient groups.

Although the generalization capabilities of deep neural networks have been hailed tremendously in the past - and often rightfully so - they might produce a false sense of security in users for applications with critical outcomes. This is why we advocate to put into context results like those of \citet{ovadia2019can} or \citet{smith2018understanding}, which suggest that some of these models are relatively robust under covariate shift. 
Such robustness was only shown in a limited circumstances and does not imply the ability to \emph{detect} covariate shift, as proven in this work. 
Our results should serve as a cautionary tale, since a strong belief in robustness to OOD coupled with a failure in OOD detection can cause serious problems in practice. We illustrate this with an example.
Consider the emergency/urgent admission group in MIMIC-III. The models are barely as good as random guessing when trying to identify the group as OOD. Moreover, the performance on mortality prediction for this group is significantly worse than performance on general population or elective admissions (mortality prediction AUC-ROC of $.71-.74$, see Figure \ref{fig:ood-both-mortality} in the appendix). Thus, in a scenario in which this group is OOD, the models in question would confidently give poor predictions. Such a discrepancy would take a considerable time to detect in practice -- several OOD patients would have to be treated before realizing that the performance of the model is declining -- time in which the model's mispredictions could already produce harm.

As advocated by \citet{thagaard2020can}, a more mindful application of uncertainty estimation techniques is required, especially in safety-critical domains such as health care. This work provides a suite of tests for OOD detection to benchmark uncertainty estimation techniques on two large public medical data sets; more investigation will be necessary to enable safe applications.

\bibliography{bibliography}


\appendix

\section{Uncertainty metrics}\label{app:uncertainty-metrics}

In the following, we denote a neural network as a function $p_{\btheta}: \mathbb{R}^D \rightarrow [0, 1]$ parameterized by a parameter vector $\btheta$. The baseline for OOD detection  introduced by \citet{hendrycks2017baseline} involves reporting the highest softmax probability across all classes:

\[
    y_{\text{max}} = \max_{c \in C}p_{\btheta}(y=c|\vec{x}_i)
\]

The underlying intuition is that the model would predict more uniform distributions for samples with higher uncertainty, therefore producing a lower $y_{\text{max}}$ score.
The uncertainty estimation techniques we consider in this work try to approximate the uncertainty by retrieving a set of $K$ predictions $p_{\btheta}^{(k)}(y|\vec{x}_i)$ for a sample $\vec{x}_i$ from either independently trained models in an ensemble, multiple forward passes with different dropout masks or models with distinct weights sampled from a (variational) posterior.
These can then be aggregated in different manners. One of the simplest ways in a binary classification setting is to compute the standard deviation of the positive class:

\[
    \sigma_\mathcal{P} = \Expect[\Big]{\mathcal{P}}{p_{\btheta}(y=1|\vec{x}_i) - \Expect[\big]{\mathcal{P}}{p_{\btheta}(y=1|\vec{x}_i)}}
\]

where $\Expect[\big]{\mathcal{P}}{p_{\btheta}(y|\vec{x}_i)}$ is commonly evaluated as a Monte-Carlo estimate using a set of model weights $\mathcal{P} = \{{\btheta_1}, \ldots, {\btheta_K}\}$ s.t.
$\frac{1}{K}\sum_{k=1}^K p_{\btheta}^{(k)}(y|\vec{x}_i) \approx \Expect[\big]{\mathcal{P}}{p_{\btheta}(y|\vec{x}_i)} $. Thus, a high standard deviation can be interpreted as a large degree of disagreement or uncertainty between classifiers. Another common metric is the predictive entropy \citep{gal2016uncertainty}:

\[
    \tilde{\mathbb{H}}[p_{\btheta}(y|\vec{x}_i)] = \mathbb{H}\Big[\Expect[\big]{\mathcal{P}}{p_{\btheta}(y|\vec{x}_i)}\Big]
\]

where $\mathbb{H}$ denotes Shannon entropy. As the entropy is computed on the averaged aggregated predictions, the entropy is low when probability mass is distributed uniformly across all classes, implying that all predictions where not able to single out a clear candidate class. Conversely, entropy is high when all classifiers accumulate mass on a single class, showing a low degree of uncertainty about the current sample. Lastly, we also consider mutual information approximated as in \citet{smith2018understanding}:

\[
    \mathbb{I}(y, {\btheta}| \vec{x}_i) \approx \mathbb{H}\Big[\overline{p}_{\btheta}(y|\vec{x}_i)\Big] - \Expect[\Big]{\mathcal{P}}{\mathbb{H}\big[p_{\btheta}(y|\vec{x}_i)\big]}
\]

where $\overline{p}_{\btheta}(y|\vec{x}_i) = \Expect[\big]{\mathcal{P}}{p_{\btheta}(y|\vec{x}_i)}$. Intuitively, this is supposed to measure the information gain about the ideal model parameters by receiving a label $y$. If the possible gain is low, that means that current parameters are close to the optimal ones, demonstrating a low uncertainty.
While other metrics capture both aleatoric and epistemic uncertainty of the model, approximate mutual information is exclusively concerned with the latter. 

\section{MIMIC-III and eICU Clinical Variables}\label{app:mimic-eicu-features}

\begin{table*}[ht!]
\floatconts
    {tab:variable-names}
    {\caption{Overview of all clinical variables used for the mortality prediction task, including their names in the MIMIC-III and the eICU data set.}}%
    {%
    \resizebox{2\columnwidth}{!}{%
    \begin{tabular}{lll}
        \toprule
        Description & MIMIC & eICU \\
        \midrule
        Diastolic Blood Pressure & \tt{Diastolic blood pressure} & \tt{Invasive BP Diastolic} \\
        Systolic Blood Pressure & \tt{Systolic blood pressure}  & \tt{Invasive BP Systolic} \\
        Fraction of inspired oxygen & \tt{Fraction inspired oxygen} & \tt{FiO2} \\ 
        Glasgow coma scale (verbal) & \tt{Glasgow coma scale verbal response} & \tt{Verbal} \\
        Glasgow coma scale (eyes) & \tt{Glasgow coma scale eyes opening} & \tt{Eyes} \\
        Glasgow coma scale (motor functions) & \tt{Glasgow coma scale motor response} & \tt{Motor} \\
        Glasgow coma scale (total) & \tt{Glasgow coma scale total} & \tt{GSC Total} \\
        Blood glucose level & \tt{Glucose} & \tt{glucose} \\
        Heart rate & \tt{Heart Rate} & \tt{Heart Rate} \\
        Mean arterial pressure & \tt{Mean blood pressure} & \tt{MAP (mmHg)} \\
        Blood oxygen saturation & \tt{Oxygen saturation} & \tt{O2 Saturation} \\
        Respiratory rate & \tt{Respiratory rate} & \tt{Respiratory Rate} \\
        Body temperature & \tt{Temperature} & \tt{Temperature (C)} \\
        Blood pH value & \tt{pH} & \tt{pH} \\
        \bottomrule
    \end{tabular}%
    }
    }
\end{table*}

An overview of the used clinical features and their corresponding names in MIMIC-III and eICU is given in Table \ref{tab:variable-names}. In all experiments, only features engineered based on those variables were used.

\section{Hyperparameter Search}\label{app:hyperparams}

For the hyperparameter search, a random search procedure \citep{bergstra2012random} is employed, sampling hyperparameters for every run from scratch from the options / distributions listed in Table \ref{tab:hyperparameters-search-space}. Per data set, we test $40$ runs for the autoencoder, vanilla neural network and MC Dropout model, while BBB receives a total of $60$ runs each. We also test all configurations for \texttt{C} for logistic regression on both data sets. The same hyperparameters found for the neural network model is then also used for its platt-scaling variant and all the ensemble models. The prior scale parameter $\lambda$ for the anchored ensemble \citep{pearce2020uncertainty} is set to $\sqrt{2 / k}$, where $k$ refers to the number of rows of a weight matrix / bias. The number of components for PPCA was set to $15$ for both data sets to be comparable with the autoencoder. All models were trained exclusively on CPU.

The best hyperparameters found (excluding the cases above where the same hyperparameters were reused for similar models) are listed in Table \ref{tab:best_hyperparameters}.

\begin{table}[ht!]
    \floatconts
        {tab:best_hyperparameters}%
        {\caption{Best hyperparameters found on the MIMIC-III and eICU data set.}}%
        {
            \resizebox{\columnwidth}{!}{%
            \begin{tabular}{llll}
                \toprule
                Model & Hyperparameter & Value MIMIC-III & Value eICU \\
                \midrule
                \texttt{AE} & \texttt{hidden\_sizes} & $75$ & $100$ \\
                \texttt{AE} & \texttt{latent\_dim} & $15$ & $15$ \\
                \texttt{AE} & \texttt{lr} & $0.006897$ & $0.005216$ \\
                \texttt{LogReg} & \texttt{C} & $10$ & $1000$ \\
                \texttt{NN} & \texttt{hidden\_sizes} & $30$ & $75$ \\
                \texttt{NN} & \texttt{dropout\_rate} & $0.157483$ & $0.381918$ \\
                \texttt{NN} & \texttt{lr} & $0.000538$ & $0.000904$ \\
                \texttt{MCDropout} & \texttt{hidden\_sizes} & $50$ & $50$ \\
                \texttt{MCDropout} & \texttt{dropout\_rate} & $0.333312$ & $0.333312$ \\
                \texttt{MCDropout} & \texttt{lr} & $0.000526$ & $0.000526$ \\
                \texttt{BBB} & \texttt{hidden\_sizes} & $[25, 25, 25]$ & $[30, 30]$ \\
                \texttt{BBB} & \texttt{dropout\_rate} & $0.177533$ & $0.038759$ \\
                \texttt{BBB} & \texttt{lr} & $0.002418$ & $0.002287$ \\
                \texttt{BBB} & \texttt{posterior\_mu\_init} & $0.22187$ & $0.518821$ \\
                \texttt{BBB} & \texttt{posterior\_rho\_init} & $-5.982621$ & $-4.475038$ \\
                \texttt{BBB} & \texttt{prior\_pi} & $0.233419$ & $0.858602$ \\
                \texttt{BBB} & \texttt{prior\_sigma\_1} & $0.740818$ & $0.904837$ \\
                \texttt{BBB} & \texttt{prior\_sigma\_2} & $0.606531$ & $0.67032$ \\
                \bottomrule
            \end{tabular}%
            }
        }
\end{table}

\begin{table*}[ht!]
    \floatconts
        {tab:hyperparameters-search-space}%
        {\caption{Distributions or options that hyperparameters were sampled from during the random hyperparameter search. $^{(*)}$For more information about these hyperparameters the reader is referred to the work of \citet{blundell2015weight}.}}%
        {
            \resizebox{2\columnwidth}{!}{%
                \begin{tabular}{llll}
                \toprule
                Hyperparameter & Description & Used in & Chosen from \\
                \midrule
                \tt{C} & Inverse $l_2$ regularization weight & \tt{LogReg} & $\{10, 100, 1000, 10000\}$ \\
                \tt{hidden\_sizes} & Number / size of hidden layers & All \texttt{NN} models, \tt{AE} & 1-4 layers of either $25$, $30$, $50$, $75$, $100$ units \\
                \tt{latent\_dim} & Dimensionality of latent space & \tt{AE} & $\{5, 10, 15, 20\}$ \\
                \tt{lr} & Learning rate & All models except \tt{PPCA} &  $\mathcal{U}(\log(10^{-4}), \log(0.1))$ \\
                \tt{dropout\_rate} & Dropout rate & All \texttt{NN} models & $\mathcal{U}(0, 0.5)$ \\ 
                \tt{posterior\_rho\_init} & Variance parameter of weight posterior$^{(*)}$ & \tt{BBB} & $\mathcal{U}(-8, -2)$ \\
                \tt{posterior\_mu\_init} & Mean parameter of weight posterior$^{(*)}$ & \tt{BBB} & $\mathcal{U}(-0.6, 0.6)$ \\
                \tt{prior\_pi} & Mixture component of prior$^{(*)}$ & \tt{BBB} & $\mathcal{U}(\exp(0.1), \exp(0.9))$\\
                \tt{prior\_sigma\_1} & Variance of prior mixture component 1$^{(*)}$ & \tt{BBB} & $\mathcal{U}(\exp(-0.8), \exp(0.1))$\\
                \tt{prior\_sigma\_2} & Variance of prior mixture component 2$^{(*)}$ & \tt{BBB} & $\mathcal{U}(\exp(-0.8), \exp(0.1))$\\
                \bottomrule
            \end{tabular}%
            }
        }
\end{table*}

\section{Additional Results}\label{app:additional_plots}

This section contains additional results and plots that were omitted due to spatial constraints.

\begin{figure}[ht]
    \floatconts
        {fig:eicu_elective_admissions}%
        {\caption{Plot of patients by elective admissions in eICU, projected into a two-dimensional space using t-SNE. 
        }}%
        {\includegraphics[width=\columnwidth]{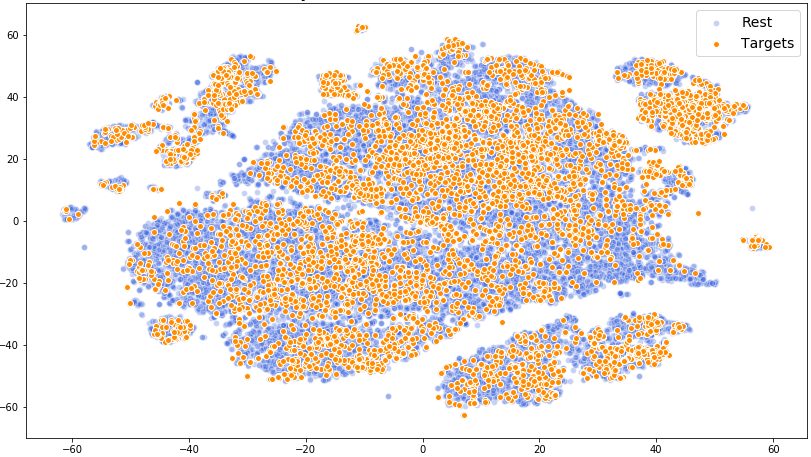}}
\end{figure}

\begin{figure}[ht]
    \floatconts
        {fig:eicu-mimic-manifold}%
        {\caption{Joint plot of the eICU and MIMIC-III manifolds, projected into a two-dimensional space using t-SNE.}}%
        {\includegraphics[width=\columnwidth]{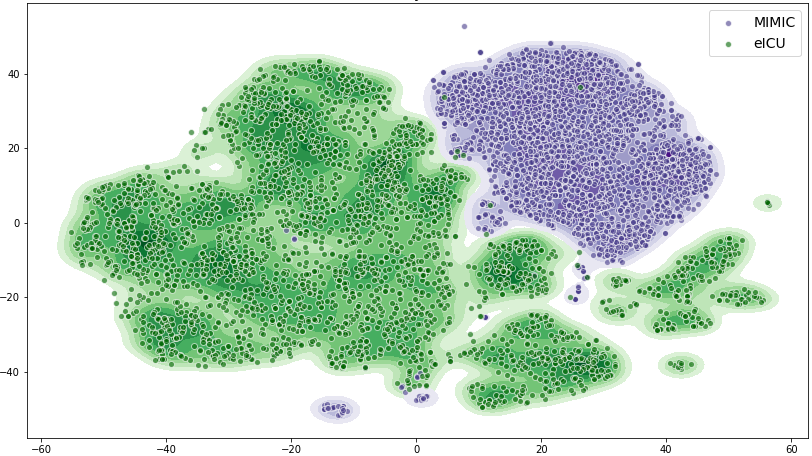}}
\end{figure}

\begin{table}[ht]
    \floatconts
        {tab:da-mortality}%
        {\caption{Mortality prediction AUC-ROC on the domain adaptation task. Results are taken over $n=5$ runs.}}%
        {
            \resizebox{\columnwidth}{!}{%
            \begin{tabular}{lll}
                \toprule
                Model &             eICU $\rightarrow$ MIMIC &  MIMIC $\rightarrow$ eICU \\
                \midrule
                \texttt{AnchoredNNEnsemble}     &  $0.91 \pm 0.00$ &  $0.88 \pm 0.00$ \\
        \texttt{BBB}                    &  $0.66 \pm 0.02$ &  $0.67 \pm 0.03$ \\
        \texttt{BootstrappedNNEnsemble} &  $0.87 \pm 0.00$ &  $0.87 \pm 0.00$ \\
        \texttt{LogReg}                 &  $0.84 \pm 0.00$ &  $0.86 \pm 0.00$ \\
        \texttt{MCDropout}              &  $0.87 \pm 0.00$ &  $0.87 \pm 0.00$ \\
        \texttt{NNEnsemble}             &  $0.87 \pm 0.00$ &  $0.87 \pm 0.00$ \\
        \texttt{NN}                     &  $0.87 \pm 0.00$ &  $0.87 \pm 0.00$ \\
        \texttt{PlattScalingNN}         &  $0.87 \pm 0.00$ &  $0.87 \pm 0.00$ \\
                \bottomrule
                \end{tabular}%
            }
    }
\end{table}

\begin{figure*}[ht]
    \floatconts
        {fig:ood_eicu}%
        {\caption{OOD detection AUC-ROC for different models and metrics given pre-defined groups in the eICU data set. \emph{size} denoted the relative size of the OOD group compared to the full training set, \emph{diff} the percentage of OOD features that were statistically significant compared to the remaining training set, using a Welch's t-test and $p < 0.01$. Results are averaged over $n=5$ runs. }}%
        {\includegraphics[width=2\columnwidth]{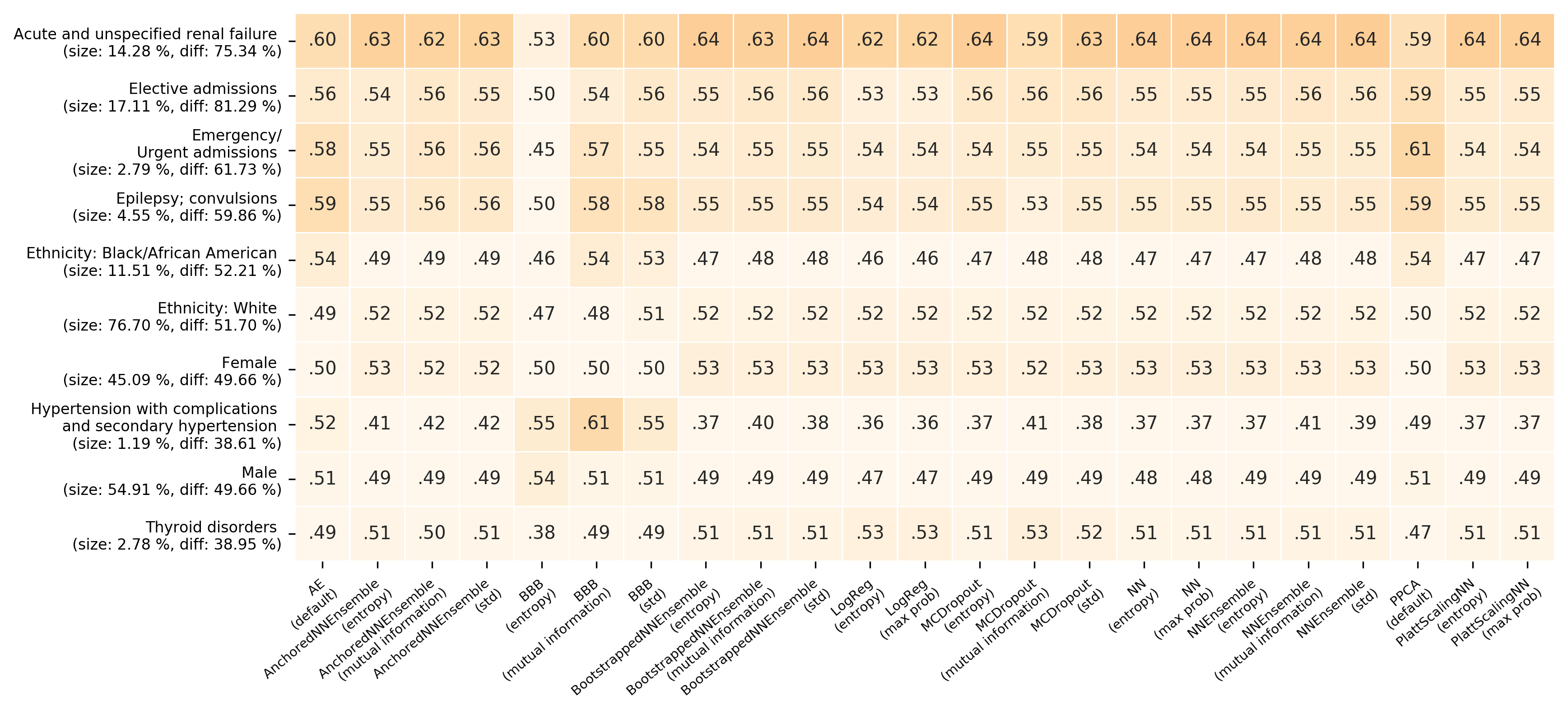}}
\end{figure*}

\begin{figure*}[h]
    \floatconts
        {fig:perturbation_mimic}%
        {\caption{Perturbation experiment results for MIMIC-III measured via the OOD detection AUC-ROC. Scales are given on the y-axis, tested models and metrics on the x-axis. Results are averaged over $n=100$ different, randomly selected perturbed features. }}%
        {\includegraphics[width=2\columnwidth]{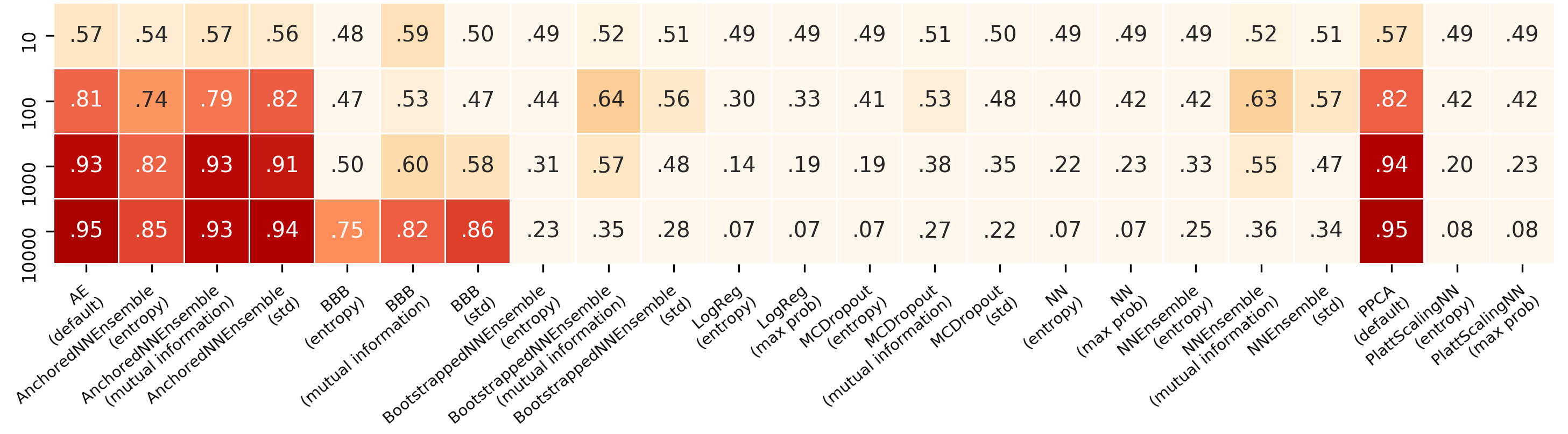}}
\end{figure*}

\begin{figure*}[ht]
    \floatconts
        {fig:compare-ensembles}%
        {\caption{Comparing the predictive entropy scores across a feature space for a multi-class toy example for a single neural discriminator (left), an ensemble of discriminators (center) and an anchored ensemble (right; \citet{pearce2020uncertainty}). While the single discriminator produces open regions with low uncertainty even far away from the training data, this effect is somewhat mitigated by ensembling, where overlapping slightly different decision boundaries create regions of higher uncertainty. In the case of the anchored ensemble, this effect is even more pronounced due to the diversity of ensemble members.}}%
        { 
            \begin{tabular}{ccc}
            \includegraphics[width=0.6\columnwidth]{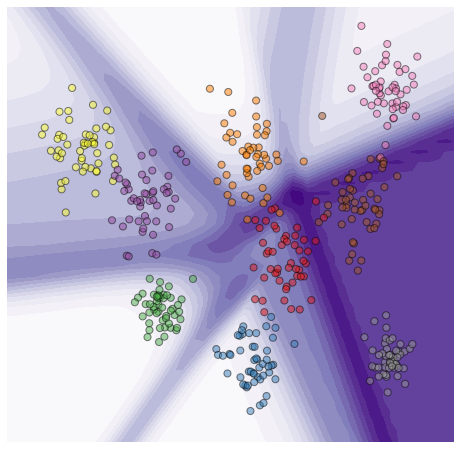} &
            \includegraphics[width=0.6\columnwidth]{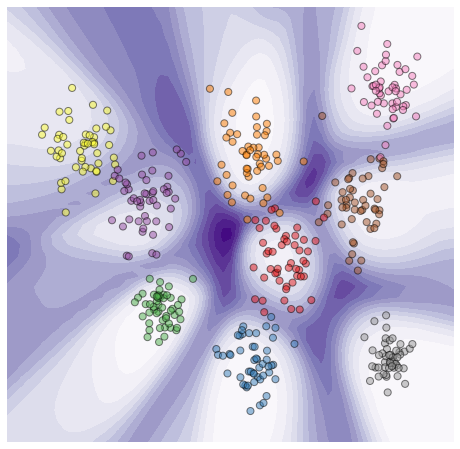} & \includegraphics[width=0.6\columnwidth]{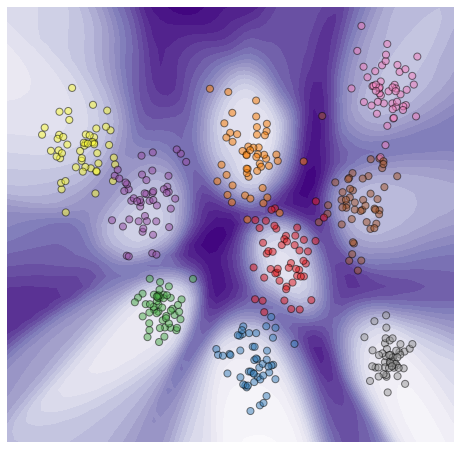} \\ 
            \\
            \end{tabular}
        }
\end{figure*}

\begin{figure*}[ht]
    \floatconts
        {fig:ood-both-mortality}%
        {\caption{Mortality prediction AUC-ROC scores for different OOD groups of the MIMIC-III (left) and eICU (right) data set. Results are averaged over $n=5$ runs.}}%
        {
            \begin{tabular}{cc}
                \includegraphics[width=0.975\columnwidth]{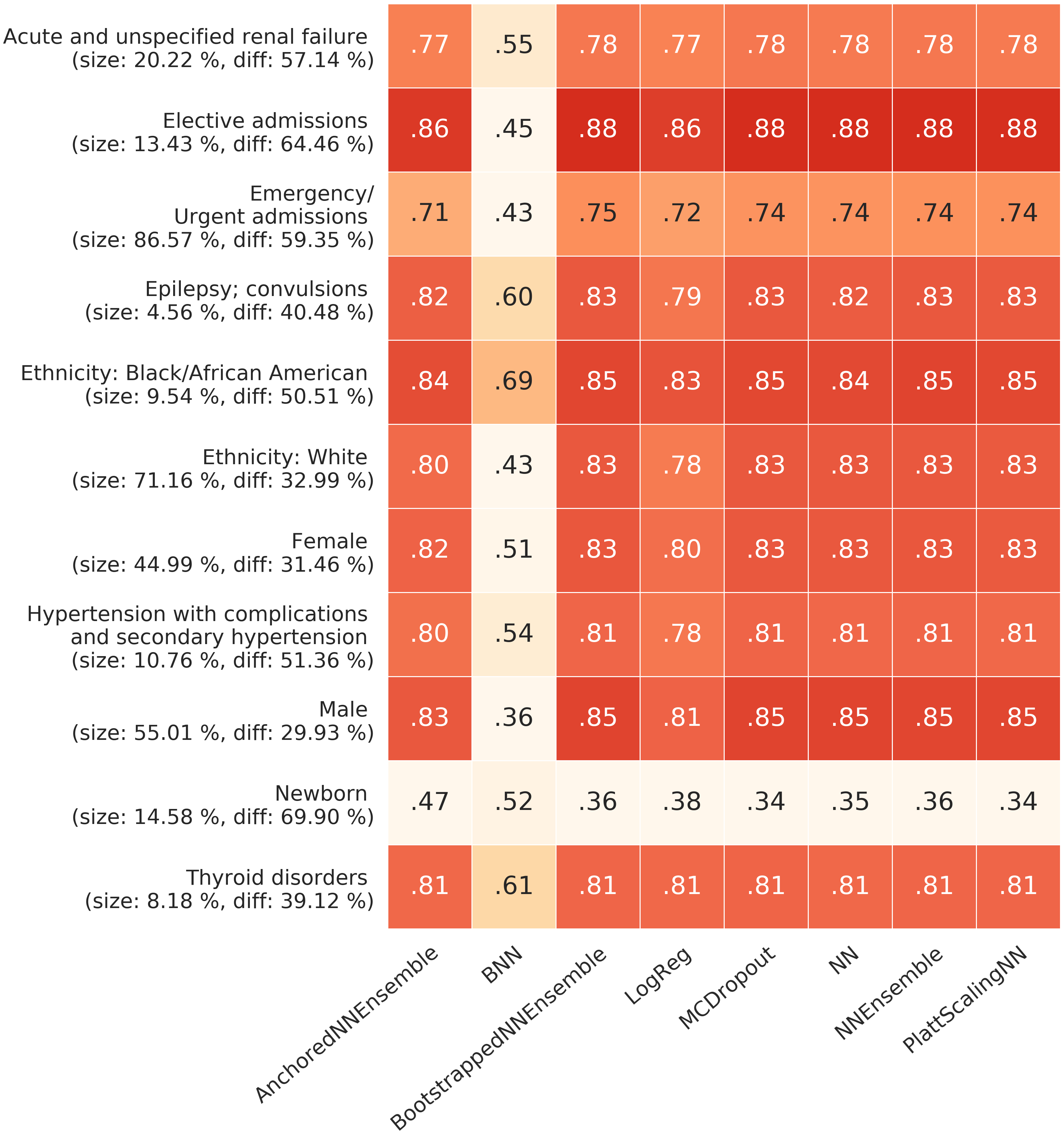} & \includegraphics[width=1.025\columnwidth]{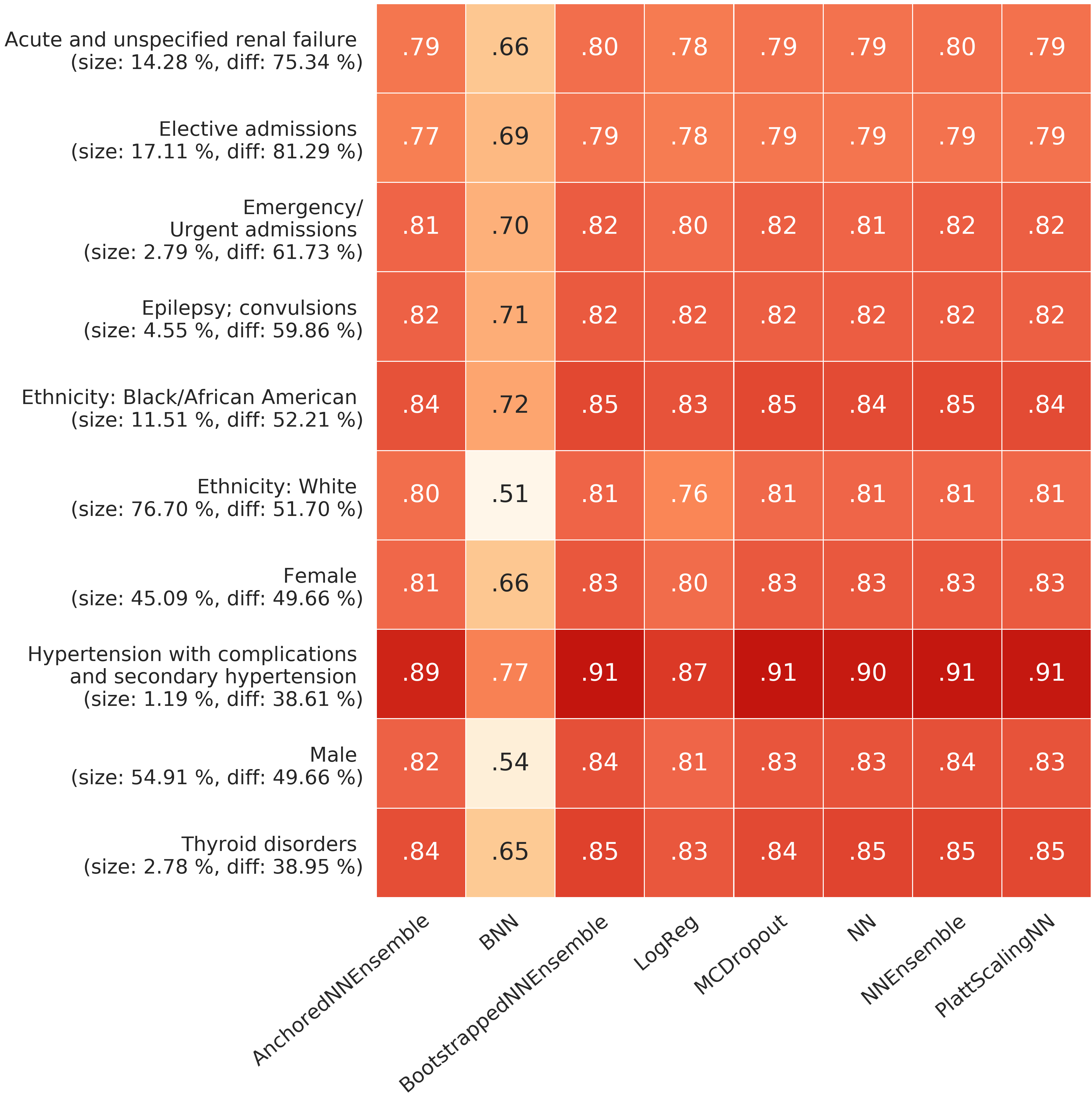} \\
            \end{tabular}
        }
\end{figure*}

\end{document}